\documentclass{article}

\usepackage[final]{nips_2018}

\usepackage[utf8]{inputenc} 
\usepackage[T1]{fontenc}    
\usepackage{hyperref}       
\usepackage{url}            
\usepackage{booktabs}       
\usepackage{amsfonts}       
\usepackage{nicefrac}       
\usepackage{microtype}      
\usepackage{bm}             
\usepackage{pgf}            
\usepackage{graphicx}
\usepackage{subcaption}

\usepackage{amsmath,amsfonts,bm,xspace}
\usepackage{array}

\newcommand{\half}{\frac{1}{2}}

\newcommand{\gcn}{GCN\xspace}
\newcommand{\gat}{GAT\xspace}
\newcommand{\monet}{MoNet\xspace}
\newcommand{\gsmean}{GS-mean\xspace}
\newcommand{\gsmaxpool}{GS-maxpool\xspace}
\newcommand{\gsmeanpool}{GS-meanpool\xspace}
\newcommand{\mlp}{MLP\xspace}
\newcommand{\logr}{LogReg\xspace}
\newcommand{\lp}{LabelProp\xspace}
\newcommand{\nllp}{LabelProp NL\xspace}

\newcommand{\cora}{CORA\xspace}
\newcommand{\citeseer}{CiteSeer\xspace}
\newcommand{\corabig}{CORA-Full\xspace}
\newcommand{\pubmed}{PubMed\xspace}
\newcommand{\magcs}{Coauthor CS\xspace}
\newcommand{\magph}{Coauthor Physics\xspace}
\newcommand{\amcomp}{Amazon Computers\xspace}
\newcommand{\amphoto}{Amazon Photo\xspace}

\newcommand{\PreserveBackslash}[1]{\let\temp=\\#1\let\\=\temp}
\newcolumntype{C}[1]{>{\PreserveBackslash\centering}p{#1}}
\newcolumntype{R}[1]{>{\PreserveBackslash\raggedleft}p{#1}}
\newcolumntype{L}[1]{>{\PreserveBackslash\raggedright}p{#1}}

\newcommand{\newterm}[1]{{\bf #1}}

\def\Tableref#1{Table~\ref{#1}}

\def\Figref#1{Figure~\ref{#1}}

\def\Secref#1{Sec.~\ref{#1}}

\def\eqref#1{equation~\ref{#1}}

\def\1{\bm{1}}

\DeclareMathAlphabet{\mathsfit}{\encodingdefault}{\sfdefault}{m}{sl}
\SetMathAlphabet{\mathsfit}{bold}{\encodingdefault}{\sfdefault}{bx}{n}

\title{Pitfalls of Graph Neural Network Evaluation}

\author{
  Oleksandr Shchur\thanks{Equal contribution}, Maximilian Mumme\footnotemark[1],
  Aleksandar Bojchevski, Stephan G\"unnemann \\
  Technical University of Munich, Germany\\
  \texttt{\{shchur,mumme,a.bojchevski,guennemann\}@in.tum.de} \\
}

\begin{document}

\maketitle

\begin{abstract}
  Semi-supervised node classification in graphs is a fundamental problem in graph mining, and the recently proposed
  graph neural networks (GNNs) have achieved unparalleled results on this task.
  Due to their massive success, GNNs have attracted a lot of attention, and many novel architectures have been put forward.
  In this paper we show that existing evaluation strategies for GNN models have serious shortcomings.
  We show that using the same train/validation/test splits of the same datasets,
  as well as making significant changes to the training procedure (e.g. early stopping criteria)
  precludes a fair comparison of different architectures.
  We perform a thorough empirical evaluation of four prominent GNN models and show that considering different splits of the data leads to dramatically different rankings of models.
  Even more importantly,
  our findings suggest that simpler GNN architectures are able to outperform the more sophisticated
  ones if the hyperparameters and the training procedure are tuned fairly for all models.
\end{abstract}


\section{Introduction}
\label{sec:introduction}
Semi-supervised node classification in graphs is a classic problem in graph mining with applications ranging from e-commerce to computational biology.
The recently proposed graph neural network architectures have achieved unprecedented results on this task and significantly advanced the state of the art.
Despite their massive success, we cannot accurately judge the progress being made 
due to certain problematic aspects of the empirical evaluation procedures.
We can partially attribute this to the practice of replicating the experimental settings from earlier works, since they are perceived as standard.
First, a number of proposed models have all been tested exclusively on the same train/validation/test splits of the same three datasets (\cora, \citeseer and \pubmed) from \cite{planetoid}.
Such experimental setup favors the model that overfits the most and defeats the main purpose of using a train/validation/test split ---
finding the model with the best generalization properties \citep{friedman2001elements}.
Second, when evaluating performance of a new model, people often use a training procedure that is rather different from the one used for the baselines.
This makes it difficult to identify whether the improved performance comes from (a) a superior architecture of the new model,
or (b) a better-tuned training procedure and / or hyperparameter configuration that unfairly benefits the new model \citep{lipton2018troubling}.

\setcounter{footnote}{0}
In this paper we address these issues and perform a thorough experimental evaluation of four prominent GNN architectures on the transductive semi-supervised node classification task.
We implement the four models -- GCN \citep{gcn}, MoNet \citep{monet}, GraphSage \citep{graphsage} and GAT \citep{gat} -- within the same framework.\footnote{Code is available at \url{https://www.kdd.in.tum.de/gnn-benchmark}}
In our evaluation we focus on two aspects:
We use a standardized training and hyperparameter selection procedure for all models.
In such a setting, the differences in performance can with high certainty be attributed to the differences in model architectures, not other factors.
Second, we perform~experiments on four well-known citation network datasets, as well as introduce four new datasets for the node classification problem.
For each dataset we use 100 random train/validation/test splits and perform 20 random initializations for each split.
This setup allows us to more accurately assess the generalization performance of different models, and does not just select the model that overfits one~fixed test set.

Before we continue, we would like to make a disclaimer, that we do not believe that accuracy on benchmark datasets is the only important characteristic of a machine learning algorithm.
Developing and generalizing the theory for existing methods, establishing connections to (and adapting ideas from) other fields 
are important research directions that move the field forward.
However, thorough empirical evaluation is crucial for understanding the strengths and limitations of different models.
\section{Models}
\label{sec:models}
We consider the problem of semi-supervised transductive node classification in a graph, as defined in \cite{planetoid}.
In this paper we compare the four following popular graph neural network architectures.
\textbf{Graph Convolutional Network (GCN)}
\citep{gcn}
is one of the earlier models that works by performing a linear approximation to spectral graph convolutions.
\textbf{Mixture Model Network (MoNet)}
\citep{monet}
generalizes the GCN architecture and allows to learn adaptive convolution filters.
The authors of \textbf{Graph Attention Network (GAT)}
\citep{gat} propose an attention mechanism that allows to weigh nodes in the neighborhood differently during the aggregation step.
Lastly, \textbf{GraphSAGE} \citep{graphsage} focuses on inductive node classification, but can also be applied for transductive setting.
We consider 3 variants of the GraphSAGE model from the original paper, denoted as \textbf{\gsmean}, \textbf{\gsmeanpool} and \textbf{\gsmaxpool}.

The original papers and reference implementations of all above-mentioned models consider different training procedures
including different early stopping strategies, learning rate decay, full-batch vs. mini-batch training (a more detailed description is provided in Appendix \ref{app:differences}).
Such diverse experimental setups makes it hard to empirically identify the driver behind the improved performance \citep{lipton2018troubling}.
Thus, in our experiments we use a standardized training and hyperparameter tuning procedure for all models (more details in \Secref{sec:exp-setup}) to perform a more fair comparison.

In addition, we consider four  baseline models.
\textbf{Logistic Regression (\logr)} and \textbf{Multilayer Perceptron (\mlp)} are attribute-based models that do not consider the graph structure.
\textbf{Label Propagation (\lp)} and \textbf{Normalized Laplacian Label Propagation (\nllp)} \citep{labelprop}, on the other hand,
only consider the graph structure and ignore the node attributes.
\section{Evaluation}

\paragraph{Datasets}
\label{sec:data}
For our experiments, we used the four well-known citation network datasets: \pubmed \citep{namata2012pubmed}, \citeseer and \cora from \cite{sen2008collective},
as well as the extended version of CORA from \cite{bojchevski2018deep}, denoted as \corabig.
We also introduce four new datasets for the node classification task: \magcs, \magph, \amcomp and \amphoto.
Descriptions of these new datasets, as well as statistics for all datasets can be found in Appendix \ref{app:datasets}.
For all datasests, we treat the graphs as undirected and only consider the largest connected component.

\paragraph{Setup}
\label{sec:exp-setup}

We keep the \emph{model architectures} as they are in the original papers / reference implementations.
This includes the type and sequence of layers, choice of activation functions, placement of dropout, and choices as to where to apply $L_2$ regularization.
We also fixed the number of attention heads for \gat to 8 and the number of Gaussian kernels for \monet to 2, as proposed in the respective papers.
All the models have 2 layers (input features $\rightarrow$ hidden layer $\rightarrow$ output layer).

For a more balanced comparison, however, we use the same \emph{training procedure} for all the models.
That is, we used the same optimizer (Adam \citep{adam} with default parameters),
same initialization (weights initialized according to \cite{glorot2010understanding}, biases initialized with zeros),
no learning rate decay, same maximum number of training epochs, early stopping criterion, patience and validation frequency (display step) for all models (Appendix \ref{app:grid_search}).
We optimize all model parameters (attention weights for \gat, kernel parameters for \monet, weight matrices for all models) simultaneously.
In all cases we use full-batch training (using all nodes in the training set every epoch).

Lastly, we used the exact same strategy for \emph{hyperparameter selection} for every model.
We performed an extensive grid search for learning rate, size of the hidden layer, strength of the $L_2$ regularization, and dropout probability
(Appendix \ref{app:grid_search}).
We restricted the random search space to ensure that every model has at most the same given number of trainable parameters.
For every model, we picked the hyperparameter configuration that achieved the best average accuracy on Cora and CiteSeer datasets
(averaged over 100 train/validation/test splits and 20 random initializations for each).
The chosen best-performing configurations were used for all subsequent experiments and are listed in \Tableref{table:final_hyperparameter_configurations}.
In all cases, we use 20 labeled nodes per class as the training set, 30 nodes per class as the validation set, and the rest as the test set.
\paragraph{Results}
\label{sec:exp-results}

\begin{table}
    \begin{center}
        \begin{tabular}{l C{2cm} C{2cm} C{2cm} C{2cm}}
              & \begin{tabular}{c}\textbf{CORA}\end{tabular} & \begin{tabular}{c}\textbf{CiteSeer}\end{tabular} &
                \begin{tabular}{c}\textbf{PubMed}\end{tabular} & \begin{tabular}{c}\textbf{CORA} \\ \textbf{Full}\end{tabular} \\
            \midrule
            \textbf{\gcn} & $81.5 \pm 1.3$ & $\textbf{71.9} \pm 1.9$ & $77.8 \pm 2.9$ & $\textbf{62.2} \pm 0.6$ \\
            \textbf{\gat} & $\textbf{81.8} \pm 1.3$ & $71.4 \pm 1.9$ & $\textbf{78.7} \pm 2.3$ & $51.9 \pm 1.5$ \\
            \textbf{\monet} & $81.3 \pm 1.3$ & $71.2 \pm 2.0$ & $78.6 \pm 2.3$ & $59.8 \pm 0.8$ \\
            \textbf{\gsmean} & $79.2 \pm 7.7$ & $71.6 \pm 1.9$ & $77.4 \pm 2.2$ & $58.6 \pm 1.6$ \\
            \textbf{\gsmaxpool} & $76.6 \pm 1.9$ & $67.5 \pm 2.3$ & $76.1 \pm 2.3$ & $40.7 \pm 1.5$ \\
            \textbf{\gsmeanpool} & $77.9 \pm 2.4$ & $68.6 \pm 2.4$ & $76.5 \pm 2.4$ & $40.5 \pm 1.5$ \\
            \textbf{\mlp} & $58.2 \pm 2.1$ & $59.1 \pm 2.3$ & $70.0 \pm 2.1$ & $36.8 \pm 1.0$ \\
            \textbf{\logr} & $57.1 \pm 2.3$ & $61.0 \pm 2.2$ & $64.1 \pm 3.1$ & $40.5 \pm 0.8$ \\
            \textbf{\lp} & $74.4 \pm 2.6$ & $67.8 \pm 2.1$ & $70.5 \pm 5.3$ & $50.5 \pm 1.5$ \\
            \textbf{\nllp} & $73.9 \pm 1.6$ & $66.7 \pm 2.2$ & $72.3 \pm 2.9$ & $51.0 \pm 1.0$ \\
        \end{tabular}
    \end{center}
    \begin{center}
        \begin{tabular}{l C{2cm} C{2cm} C{2cm} C{2cm}}
              & \begin{tabular}{c}\textbf{Coauthor} \\ \textbf{CS}\end{tabular} & \begin{tabular}{c}\textbf{Coauthor} \\ \textbf{Physics}\end{tabular} &
                \begin{tabular}{c}\textbf{Amazon} \\ \textbf{Computer}\end{tabular} & \begin{tabular}{c}\textbf{Amazon} \\ \textbf{Photo}\end{tabular} \\
            \midrule
            \textbf{\gcn} & $91.1 \pm 0.5$ & $92.8 \pm 1.0$ & $82.6 \pm 2.4$ & $91.2 \pm 1.2$ \\
            \textbf{\gat} & $90.5 \pm 0.6$ & $92.5 \pm 0.9$ & $78.0 \pm 19.0$ & $85.7 \pm 20.3$ \\
            \textbf{\monet} & $90.8 \pm 0.6$ & $92.5 \pm 0.9$ & $\textbf{83.5} \pm 2.2$ & $91.2 \pm 1.3$ \\
            \textbf{\gsmean} & $\textbf{91.3} \pm 2.8$ & $\textbf{93.0} \pm 0.8$ & $82.4 \pm 1.8$ & $\textbf{91.4} \pm 1.3$ \\
            \textbf{\gsmaxpool} & $85.0 \pm 1.1$ & $90.3 \pm 1.2$ & N/A & $90.4 \pm 1.3$ \\
            \textbf{\gsmeanpool} & $89.6 \pm 0.9$ & $92.6 \pm 1.0$ & $79.9 \pm 2.3$ & $90.7 \pm 1.6$ \\
            \textbf{\mlp} & $88.3 \pm 0.7$ & $88.9 \pm 1.1$ & $44.9 \pm 5.8$ & $69.6 \pm 3.8$ \\
            \textbf{\logr} & $86.4 \pm 0.9$ & $86.7 \pm 1.5$ & $64.1 \pm 5.7$ & $73.0 \pm 6.5$ \\
            \textbf{\lp} & $73.6 \pm 3.9$ & $86.6 \pm 2.0$ & $70.8 \pm 8.1$ & $72.6 \pm 11.1$ \\
            \textbf{\nllp} & $76.7 \pm 1.4$ & $86.8 \pm 1.4$ & $75.0 \pm 2.9$ & $83.9 \pm 2.7$ \\
        \end{tabular}
    \end{center}
    \caption{Mean test set accuracy and standard deviation in percent averaged over $100$ random train/validation/test splits with $20$ random weight initializations each for all models and all datasets.
    For each dataset, the highest accuracy score is marked in \textbf{bold}.
    N/A stands for the dataset that couldn't be processed by the full-batch version of \gsmaxpool because of GPU RAM limitations.
    }
    \label{table:final_experiment_results_mean}
\end{table}

\Tableref{table:final_experiment_results_mean} shows mean accuracies (and their standard deviations\footnote{Standard deviations are not the best representation of the variance of the accuracy scores, since the scores are not normally distributed.
We still include the standard deviations to give the reader a rough idea of the variance of the results for each model.
A more accurate picture is given by the box plots in \Figref{fig:final_experiment_results_plot}.}) of all models for all 8 datasets averaged over 100 splits and 20 random initializations for each split.
There are a few observations to be made.
First, the GNN-based approaches (GCN, MoNet, GAT, GraphSAGE) significantly outperform all the baselines (MLP, LogReg, LabelProp, LabelProp NL) across all the datasets.
This matches our intuition and confirms the superiority of GNN-based approaches that combine both the structural and attribute information compared to methods considering only the attributes or only the structure.

Among the GNN approaches, there is no clear winner that dominates across all the datasets.
In fact, for 5 out of 8 datasets, scores of the 2nd and 3rd best approaches are less than 1\% away from the average score of the best-performing method.
If we were interested in comparing one model versus the rest, we could perform pairwise t-tests, as done in \cite{klicpera2018personalized}.
Since we are interested in comparing all the models to each other, we consider the relative accuracy of each model instead.
For this, we take the best accuracy score for each split of each dataset (already averaged over 20 initializations) as 100\%.
Then, the score of each model is divided by this number,
and the results for each model are averaged over all the datasets and splits.
We also rank algorithms by their performance (1 = best performance, 10 = worst), and compute the average rank across all datasets and splits for each algorithm.
The final scores are reported in \Tableref{table:relative_performance_from_means}.
We observe that \gcn is able to achieve the best performance across all models.
While this result seems surprising, similar findings have been reported in other fields. Simpler models often outperform more sophisticated ones
if hyperparameter tuning is performed equally carefully for all methods \citep{melis2017state,lucic2017gans}.
In future work, we plan to further investigate what are the specific properties of the graphs that lead to the differences in performance of the GNN models.

Another surprising finding is the relatively lower score and high variance in results obtained by \gat for the \amcomp and \amphoto datasets.
To investigate this phenomenon, we additionally visualize the accuracy scores achieved by different models on the \amphoto dataset in \Figref{fig:outliers_plot} in the appendix.
While the median scores for all GNN models are very close to each other, \gat produces extremely low scores (below 40\%) for some weight initializations.
While these outliers occur rarely (for 138 out of 2000 runs), they significantly lower the average score of \gat.

\paragraph{Effect of the train/validation/test split}
To demonstrate the effect of different train/validation/test splits on the performance, we execute the following simple experiment.
We run the 4 models on the datasets and respective splits from \citep{planetoid}.
As shown in \Tableref{table:planetoid_inverted}, \gat achieves the best scores for the \cora and \citeseer datasets, and \gcn gets the top score for \pubmed.
If we, however, consider a different random split with the same train/validation/test set sizes the ranking of models is completely different,
with \gcn being first on \cora and \citeseer, and \monet winning on \pubmed.
This shows how fragile and misleading results obtained on a single split can be.
Taking further into account that the predictions of GNNs can greatly change under small data perturbations \citep{zugner2018adversarial} clearly confirms the need for evaluation strategies based on multiple splits.

\begin{table}
    \begin{center}
    	\begin{subtable}{0.37\linewidth}
        \resizebox{\textwidth}{!}{%
        \begin{tabular}{L{2.2cm} R{0.8cm} R{1cm}}
        &  \hspace*{-4.5em} \textbf{Relative} & \hspace*{-0.5em} \textbf{Avg.}\\
        &  \hspace*{-4.5em} \textbf{accuracy} & \hspace*{-0.5em} \textbf{rank}\\
            \midrule
            \textbf{\gcn} & $99.4$ & $2.3$ \\
            \textbf{\monet} & $99.0$ & $2.7$\\
            \textbf{\gsmean} & $98.3$ & $2.7$\\
            \textbf{\gat} & $95.9$ & $3.6$\\
            \textbf{\gsmeanpool} & $93.0$ & $5.2$\\
            \textbf{\gsmaxpool} & $91.1$ & $6.4$\\
            \textbf{\nllp} & $89.3$ & $7.4$\\
            \textbf{\lp} & $86.6$ & $7.7$\\
            \textbf{\logr} & $80.6$ & $8.8$\\
            \textbf{\mlp} & $77.8$ & $8.8$\\
        \end{tabular}%
        }
	    \vspace*{+0.5em}
	    \caption{Relative accuracy and average rank.}\label{table:relative_performance_from_means}
	    \end{subtable}
        \hfill
        \begin{subtable}{0.6\linewidth}
        \resizebox{\textwidth}{!}{%
        \begin{tabular}{lccc}
        	\rule{0pt}{2em}
            \textbf{Planetoid split} & \textbf{\cora} & \textbf{\citeseer} & \textbf{\pubmed} \\
            \midrule
            \textbf{\gcn} & $81.9 \pm 0.8$ & $69.5 \pm 0.9$ & $\textbf{79.0} \pm 0.5$ \\
            \textbf{\gat} & $\textbf{82.8} \pm 0.5$ & $\textbf{71.0} \pm 0.6$ & $77.0 \pm 1.3$ \\
            \textbf{\monet} & $82.2 \pm 0.7$ & $70.0 \pm 0.6$ & $77.7 \pm 0.6$ \\
            \textbf{\gsmaxpool} & $77.4 \pm 1.0$ & $67.0 \pm 1.0$ & $76.6 \pm 0.8$ \\
            & & & \\
            \textbf{Another split} & \textbf{\cora} & \textbf{\citeseer} & \textbf{\pubmed} \\
            \midrule
            \textbf{\gcn} & $\textbf{79.0} \pm 0.7$ & $\textbf{68.6} \pm 1.1$ & $69.5 \pm 1.0$ \\
            \textbf{\gat} & $77.9 \pm 0.7$ & $67.7 \pm 1.2$ & $69.5 \pm 0.6$ \\
            \textbf{\monet} & $77.9 \pm 0.7$ & $66.8 \pm 1.3$ & $\textbf{70.7} \pm 0.5$ \\
            \textbf{\gsmaxpool} & $74.5 \pm 0.6$ & $63.1 \pm 1.2$ & $70.3 \pm 0.8$ \\
        \end{tabular}%
        }
	    \caption{Different split leads to a completely different ranking of models.}\label{table:planetoid_inverted}
	\end{subtable}
    \end{center}
    \caption{
        (a) Relative accuracy scores and ranks averaged over all datasets. See text for the definition.
        (b) Model accuracy on the Planetoid split from \cite{planetoid} and another split on the same datasets.
        Different splits lead to a completely different ranking of models.}
\end{table}
\section{Conclusion}
\label{sec:conclusion}
We have performed an empirical evaluation of four state-of-the-art GNN architectures on the node classification task.
We introduced four new attributed graph datasets, as well as open-sourced a framework that enables a fair and reproducible comparison of different GNN models.
Our results highlight the fragility of experimental setups that consider only a single train/validation/test split of the data.
We also find that, surprisingly, a simple GCN model can outperform the more sophisticated GNN architectures if the
same hyperparameter selection and training procedures are used, and the results are averaged over multiple data splits.
We hope that these results will encourage future works to use more robust evaluation procedures.

\bibliographystyle{abbrvnat}
\newpage
\section*{Acknowledgments}
This research was supported by the German Research Foundation, Emmy Noether grant GU 1409/2-1.
\bibliography{report}

\appendix
\newpage

\section{Differences in training procedures for GNN models}
\label{app:differences}
\textbf{\gcn}
\begin{itemize}
    \item Early stopping: stop optimization if the validation loss is larger than the mean of validation losses of the last 10 epochs.
    \item Full-batch training.
    \item Maximum number of epochs: 200.
    \item Train set: 20 per class; validation set: 500 nodes; test set: 1000 (as in the Planetoid split).
\end{itemize}
\textbf{\monet}
\begin{itemize}
    \item No early stopping.
    \item Full-batch training.
    \item Maximum number of epochs: 3000 for \cora, 1000 for \pubmed.
    \item Train set: 20 per class; validation set: 500 nodes; test set: 1000 (as in the Planetoid split).
    \item Alternating optimization of weight matrices and kernel parameters.
    \item Learning rate decay at predefined iterations (only for \cora).
\end{itemize}
\textbf{\gat}
\begin{itemize}
    \item Early stopping: stop optimization if neither the validation loss nor the validation accuracy improve for 100 epochs.
    \item Full-batch training.
    \item Maximum number of epochs: 100000.
    \item Train set: 20 per class; validation set: 500 nodes; test set: 1000 (as in the Planetoid split).
\end{itemize}
\textbf{GraphSAGE}
\begin{itemize}
    \item No early stopping.
    \item Mini-batch training with batch size of 512.
    \item Maximum number of epochs (each epoch consists of multiple mini-batches): 10.
\end{itemize}

\newpage
\section{Datasets description and statistics}
\label{app:datasets}

\amcomp and \amphoto are segments of the Amazon co-purchase graph \citep{mcauley2015amazon}, where nodes represent goods, edges indicate that two goods are frequently bought together,
node features are bag-of-words encoded product reviews, and class labels are given by the product category.

\magcs and \magph are co-authorship graphs based on the Microsoft Academic Graph from the KDD Cup 2016 challenge
\footnote{\url{https://kddcup2016.azurewebsites.net/}}.
Here, nodes are authors, that are connected by an edge if they co-authored a paper; node features represent paper keywords for each author's papers, and class labels indicate most active fields of study for each author.

\begin{table}[h!]
    \begin{center}
        \begin{tabular}{l r r r r r r}
                & \textbf{Classes} & \textbf{Features} & \textbf{Nodes} & \textbf{Edges} & \textbf{Label rate} & \textbf{Edge density} \\
            \midrule
            \textbf{\cora}         &            7 &          1433 &       2485 &       5069 &      0.0563 &        0.0004 \\
            \textbf{\citeseer}     &            6 &          3703 &       2110 &       3668 &      0.0569 &        0.0004 \\
            \textbf{\pubmed}       &            3 &           500 &      19717 &      44324 &      0.0030 &        0.0001 \\
            \textbf{\corabig}      &           67 &          8710 &      18703 &      62421 &      0.0745 &        0.0001 \\
            \textbf{\magcs}        &           15 &          6805 &      18333 &      81894 &      0.0164 &        0.0001 \\
            \textbf{\magph}        &            5 &          8415 &      34493 &     247962 &      0.0029 &        0.0001 \\
            \textbf{\amcomp}       &           10 &           767 &      13381 &     245778 &      0.0149 &        0.0007 \\
            \textbf{\amphoto}      &            8 &           745 &       7487 &     119043 &      0.0214 &        0.0011 \\
        \end{tabular}
    \end{center}
    \caption{Dataset statistics after standardizing the graphs, adding self-loops and removing classes with too few instances from CORA\char`_full.
    We ignore 3 classes with less than 50 nodes in \corabig dataset (since we cannot perform the 20/30/rest split for them).\\\\
    \newterm{Label rate} is the fraction of nodes in the training set.
    Since we use $20$ training instances per class this can be computed as $(\textup{\#classes} \cdot 20)\ /\ \textup{\#nodes}$.\\\\
    The \newterm{edge density} describes the fraction of all possible edges that is present in the graph and can be
    computed as $\textup{\#edges}\ /\ (\half \cdot \textup{\#nodes}^2)$.
    }
    \label{table:dataset_statistics}
\end{table}

\newpage
\section{Hyperparameter configurations and Early Stopping}
\label{app:grid_search}
Grid search was performed over the following search space:
\begin{itemize}
    \item Hidden size: \texttt{[8, 16, 32, 64]}
    \item Learning rate: \texttt{[0.001, 0.003, 0.005, 0.008, 0.01]}
    \item Dropout probability: \texttt{[0.2, 0.3, 0.4, 0.5, 0.6, 0.7, 0.8]}
    \item Attention coefficients dropout probability (only for \gat):

    \texttt{[0.2, 0.3, 0.4, 0.5, 0.6, 0.7, 0.8]}
    \item $L_2$ regularization strength: \texttt{[1e-4, 5e-4, 1e-3, 5e-3, 1e-2, 5e-2, 1e-1]}
\end{itemize}

We train for a maximum of $100$k epochs. However, the actual training time is considerably shorter since we use strict early stopping. Specifically, with our unified early stopping criterion training stops if the total validation loss (loss on the data plus regularization loss) does not improve for $50$ epochs. Once training has stopped, we reset the state of the weights to the step with the lowest validation loss.

\begin{table}[h]
    \begin{center}
        \begin{tabular}{l r r r r r r }
        & \begin{tabular}{c} \textbf{Effective}\\\textbf{hidden size} \end{tabular} & \begin{tabular}{c} \textbf{Learning}\\\textbf{rate} \end{tabular}
            & \begin{tabular}{c} \textbf{Dropout} \end{tabular}  & \begin{tabular}{c} \textbf{$L_2$ reg.}\\\textbf{strength} \end{tabular}
            & \begin{tabular}{c} \textbf{Trainable}\\\textbf{weights}\end{tabular} \\
            \midrule
            \textbf{\gcn} & $64$ & $0.01$ & $0.8$ & $0.001$ & $92K$ \\
            \textbf{\gat} & $64$ & $0.01$ & $0.6$/$0.3$ & $0.01$ & $92K$ \\
            \textbf{\monet} & $64$ & $0.003$ & $0.7$ & $0.05$ & $92K$ \\
            \textbf{\gsmean} & $32$ & $0.001$ & $0.4$ & $0.1$ & $92K$ \\
            \textbf{\gsmaxpool} & $32$/$32$ & $0.001$ & $0.3$ & $0.005$ & $94K$ \\
            \textbf{\gsmeanpool} & $32$/$8$ & $0.001$ & $0.2$ & $0.01$  & $58K$ \\
            \textbf{\mlp} & $64$ & $0.005$ & $0.8$ & $0.01$ & $92K$ \\
            \textbf{\logr} & -- & $0.1$ & -- & $0.0005$ & $10K$ \\
        \end{tabular}
    \end{center}
    \caption{Best performing hyperparameter configurations for each model chosen by grid search.
    GAT has two dropout probabilities (dropout on features / dropout on attention coefficients).
    All GraphSAGE models have additional weights for the skip connections (which effectively doubles the hidden size).
    \gsmeanpool/\gsmaxpool have two hidden sizes (hidden layer size / size of intermediary feature transformation).
    \gat uses a multi-head architecture with $8$ heads and \monet uses $2$ heads, so the hidden state is split over $8$ and $2$ heads respectively.
    }
    \label{table:final_hyperparameter_configurations}
\end{table}

\newpage
\section{Performance of different models across datasets}
\begin{figure}[h!]
    \begin{subfigure}[t]{0.45\textwidth}
        \centering
        \caption{\textbf{\cora}}
        \includegraphics{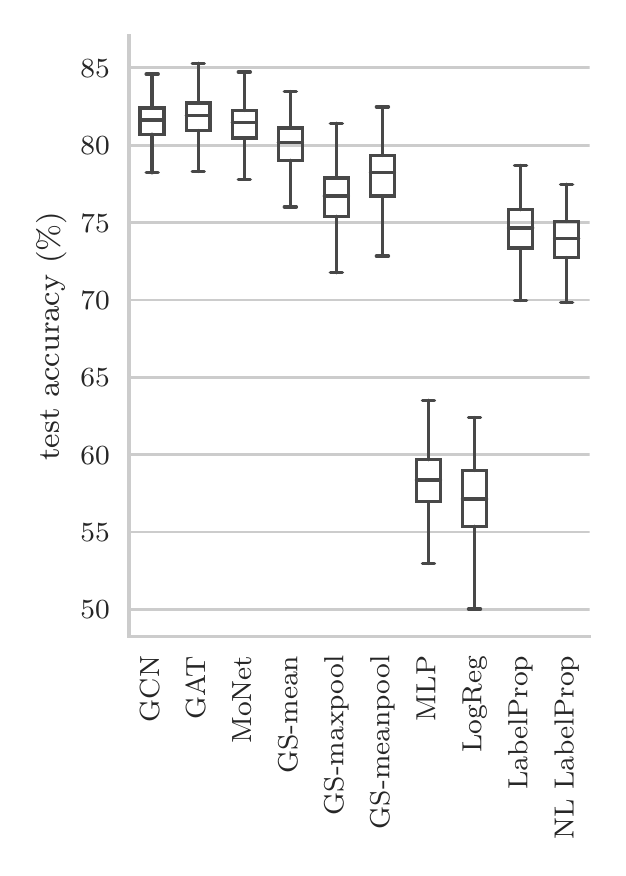}
    \end{subfigure}
    \hfill
    \begin{subfigure}[t]{0.45\textwidth}
        \centering
        \caption{\textbf{\citeseer}}
        \includegraphics{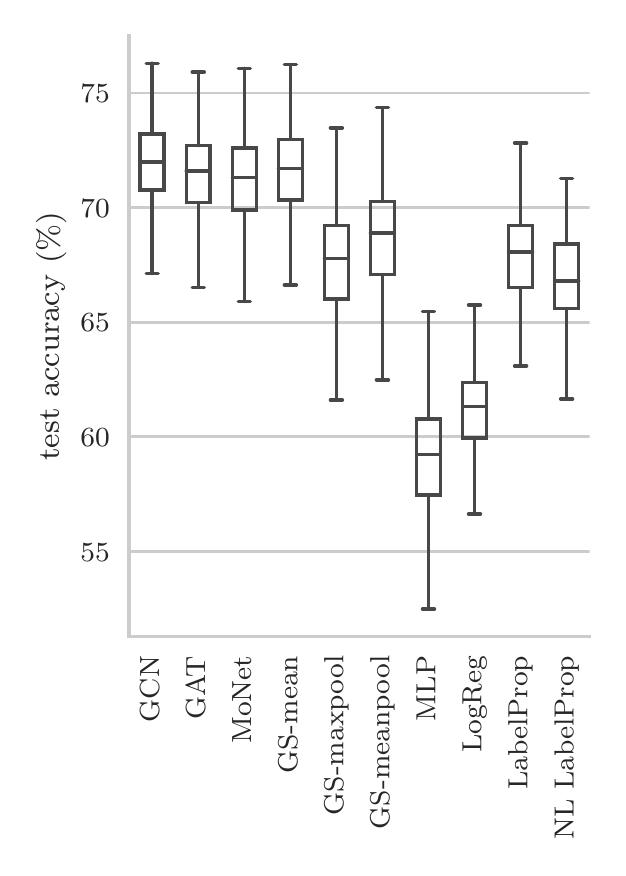}
    \end{subfigure}

    \bigskip

    \begin{subfigure}[t]{0.45\textwidth}
        \centering
        \caption{\textbf{\pubmed}}
        \includegraphics{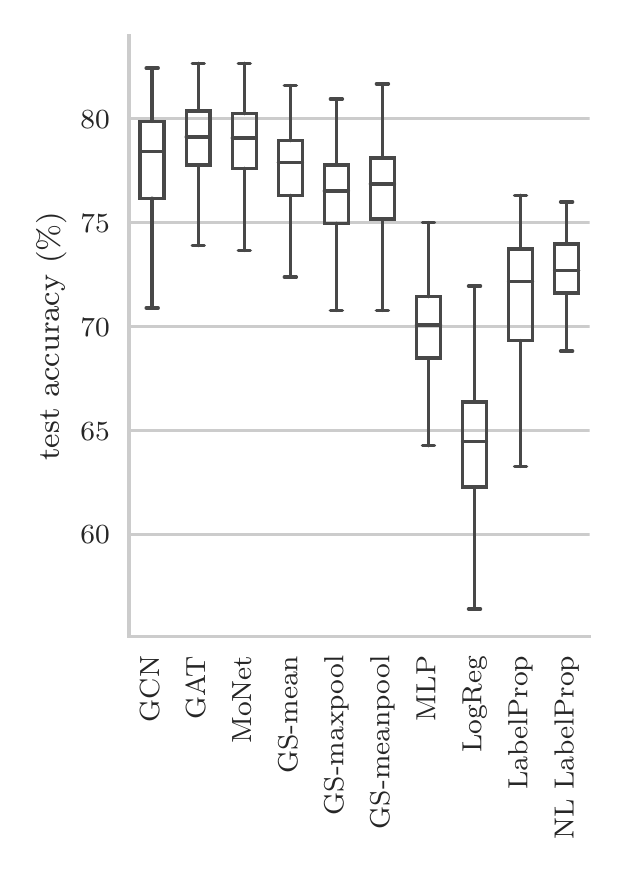}
    \end{subfigure}
    \hfill
    \begin{subfigure}[t]{0.45\textwidth}
        \centering
        \caption{\textbf{\corabig}}
        \includegraphics{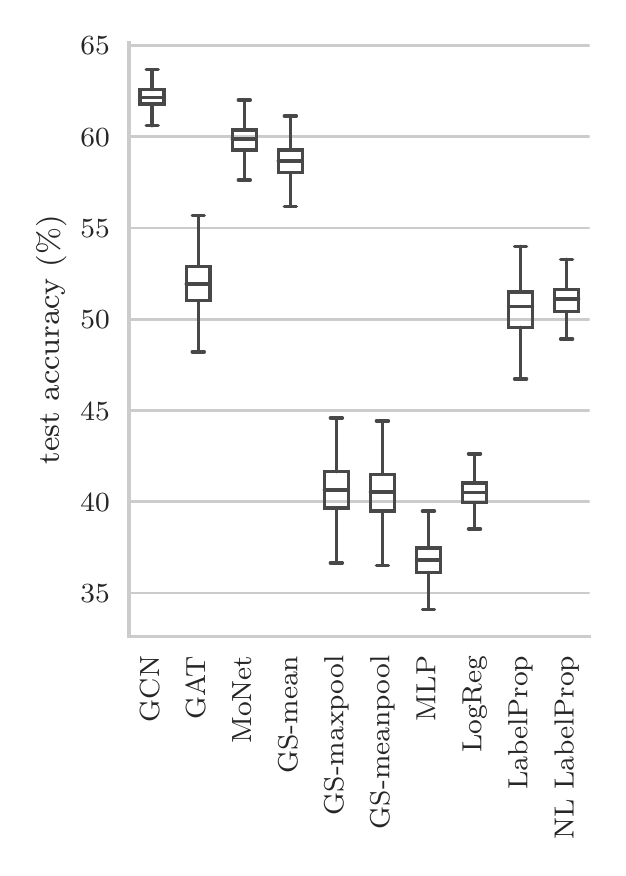}
    \end{subfigure}
\end{figure}
\begin{figure}
    \ContinuedFloat 
    \begin{subfigure}[t]{0.45\textwidth}
        \centering
        \caption{\textbf{\magcs}}
        \includegraphics{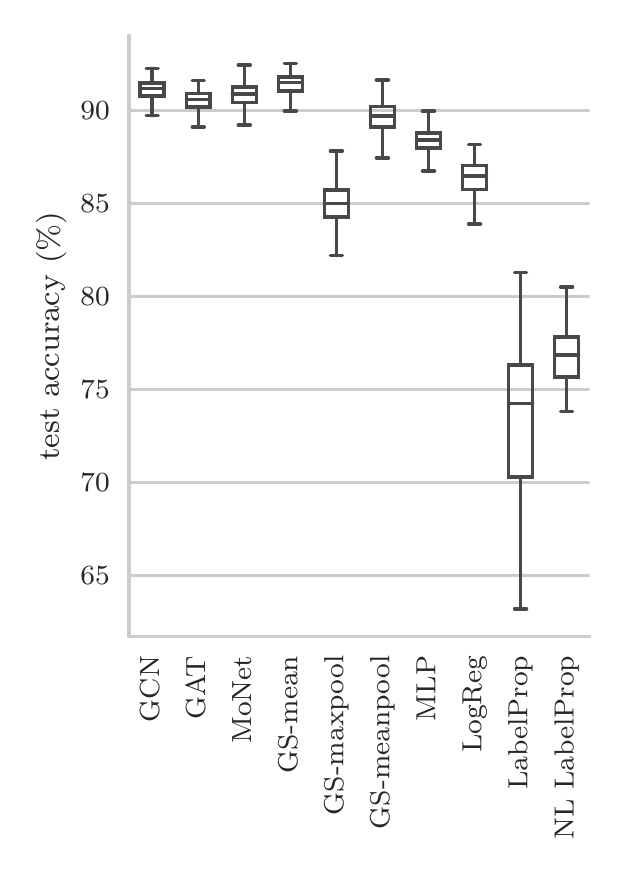}
    \end{subfigure}
    \hfill
    \begin{subfigure}[t]{0.45\textwidth}
        \centering
        \caption{\textbf{\magph}}
        \includegraphics{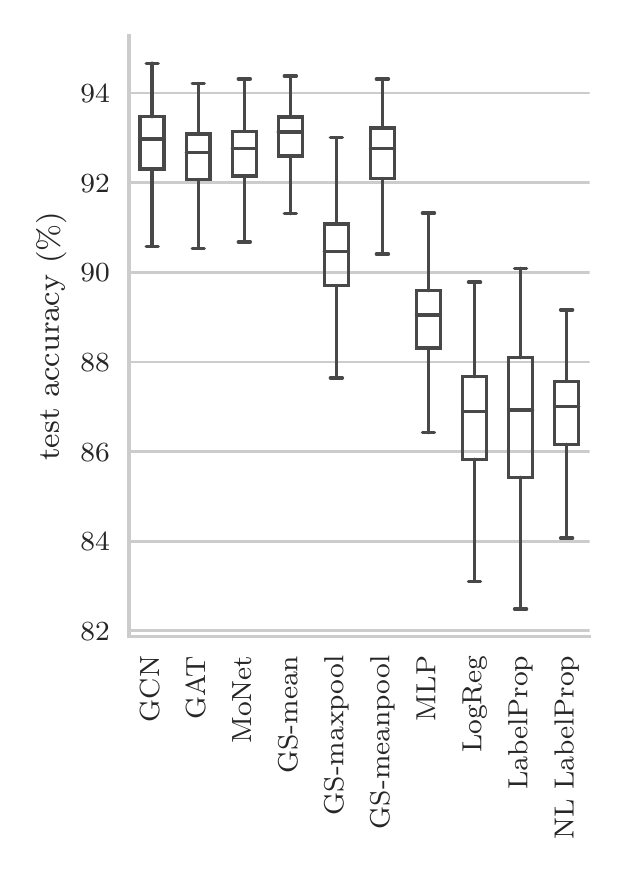}
    \end{subfigure}

    \bigskip

    \begin{subfigure}[t]{0.45\textwidth}
        \centering
        \caption{\textbf{\amcomp}}
        \includegraphics{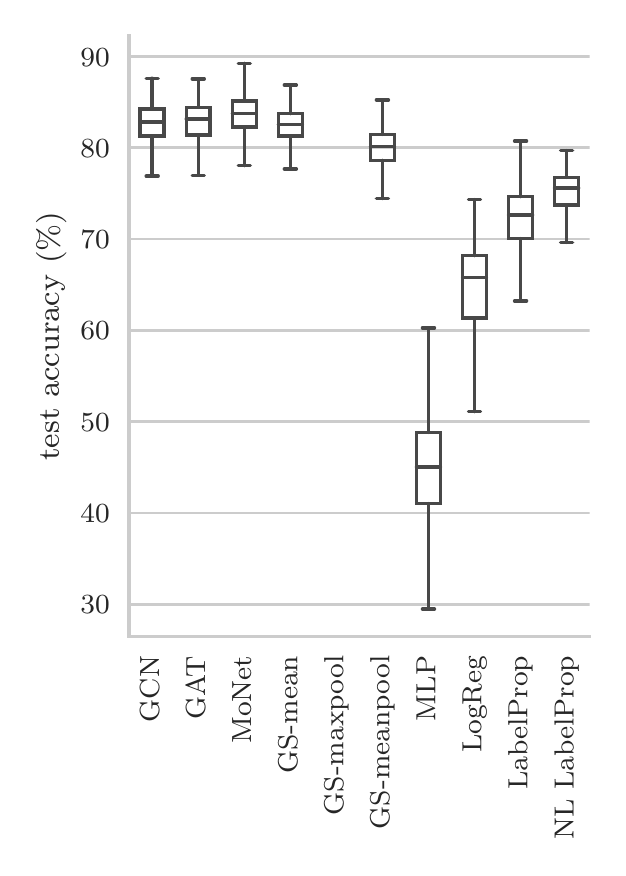}
    \end{subfigure}
    \hfill
    \begin{subfigure}[t]{0.45\textwidth}
        \centering
        \caption{\textbf{\amphoto}}
        \includegraphics{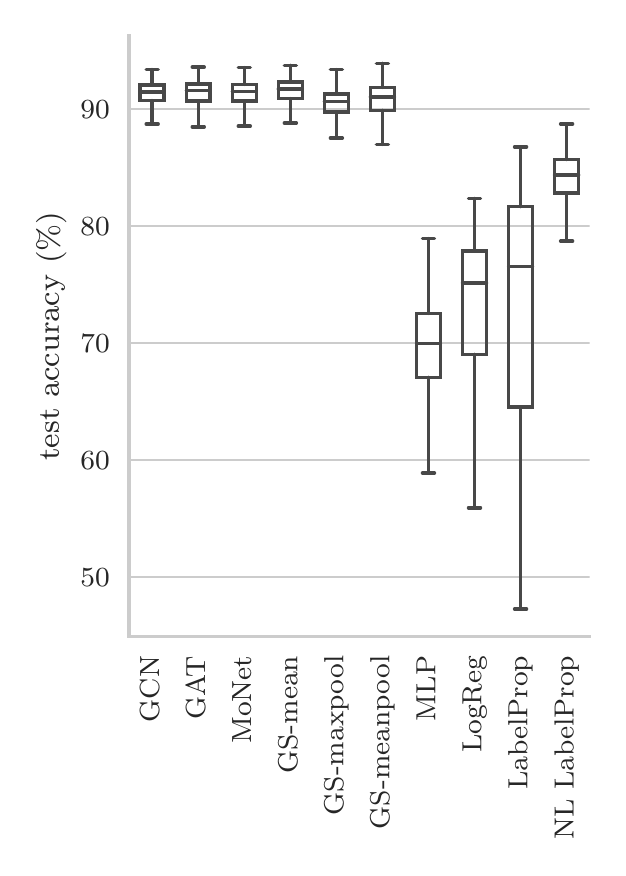}
    \end{subfigure}

    \caption{Boxplots of the test set accuracy of all models on all datasets over $100$ random train/validation/test splits with $20$ random weight initializations each.
    Note that a boxplot displays the median of the data as well as the $50\%$ quantiles.
    Note further that outliers are excluded in these plots since some models have outliers very far from the median which would shrink the resolution of the plots.
    For a plot including the outliers refer to \Figref{fig:outliers_plot}.}
\label{fig:final_experiment_results_plot}
\end{figure}
\begin{figure}
    \centering
    \includegraphics{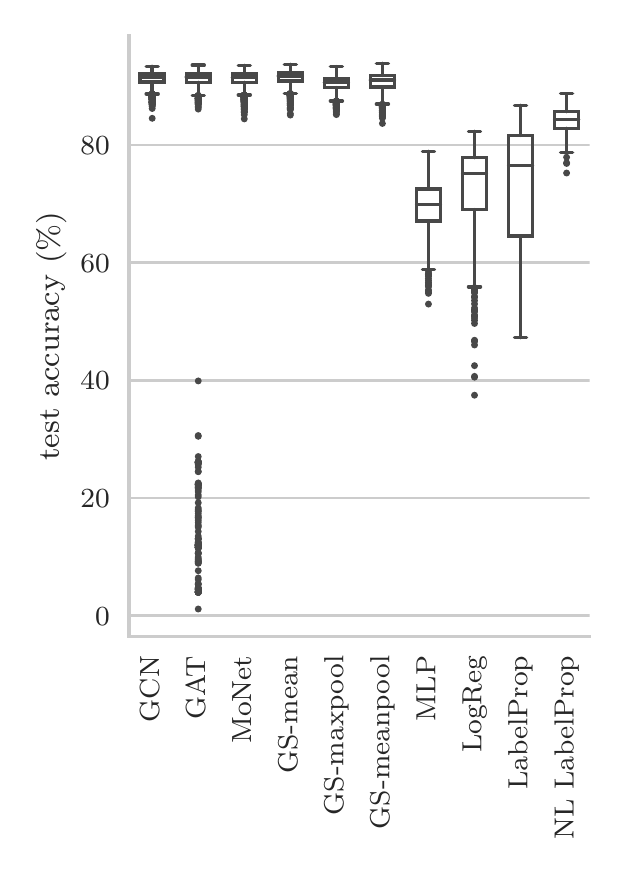}
    \caption{Boxplot showing outliers for the \textbf{\amphoto} dataset.}
\label{fig:outliers_plot}
\end{figure}

\end{document}